# Semantic Risk Scoring of Aggregated Metrics: An AI-Driven Approach for Healthcare Data Governance

1st Mohammed Omer Shakeel Ahmed
*University of Texas at Arlington*
Dallas, TX
fnu.mohammedomersha@mavs.uta.edu [0009-0001-3330-1892]

*Abstract*—Large healthcare institutions typically operate multiple business intelligence (BI) teams segmented by domain, including clinical performance, fundraising, operations, and compliance. Due to HIPAA, FERPA, and IRB restrictions, these teams face challenges in sharing patient-level data needed for analytics. To mitigate this, A metric aggregation table is proposed, which is a precomputed, privacy-compliant summary such as "Average Patient Wait Times by Department" or "Donor Conversion by Campaign." These abstractions enable decision-making without direct access to sensitive data. However, even aggregated metrics can inadvertently lead to privacy risks if constructed without rigorous safeguards. A modular AI framework is proposed that evaluates SQL-based metric definitions for potential overexposure using both semantic and syntactic features. Specifically, the system parses SQL queries into abstract syntax trees (ASTs), extracts sensitive patterns (e.g., fine-grained GROUP BY on ZIP code or gender), and encodes the logic using pretrained CodeBERT embeddings. These are fused with structural features and passed to an XGBoost classifier trained to assign risk scores. Queries that surpass the risk threshold (e.g., $> 0.85$) are flagged and returned with human-readable explanations (e.g., "Metric may overexpose sensitive groupings like gender or ZIP"). This enables proactive governance, preventing statistical disclosure before deployment. This implementation demonstrates strong potential for cross-departmental metric sharing in healthcare while maintaining compliance and auditability. The system also promotes role-based access control (RBAC), supports zero-trust data architectures, and aligns with national data modernization goals by ensuring that metric pipelines are explainable, privacy-preserving, and AI-auditable by design. Unlike prior works that rely on runtime data access to flag privacy violations, the proposed framework performs static, explainable detection at the query-level, enabling pre-execution protection and audit-readiness

## I. INTRODUCTION

In today's data-driven healthcare systems, decisions are increasingly powered by metrics summarized values that describe key trends such as patient outcomes, treatment efficiency, or fundraising performance. The explosion of data science in the modern world has increased the importance of business intelligence, or BI, in the healthcare industry. The healthcare industry generates a lot of data from patients, processes, tests, research, and operations [1]. Business analytics (BA) in healthcare research offers numerous valuable insights that can enhance patient care and hospital performance [2] Compared to data warehouses in non-healthcare settings, those in healthcare settings may focus more on privacy and security of protected health information, as well as compliance with federal and state regulations and organizational policies [3]. Metrics are especially important in healthcare, where hospitals must continuously measure performance to improve care, manage operations, and demonstrate impact. Business Intelligence (BI) tools enable health institutions to make data-informed decisions by transforming large volumes of clinical and administrative data into actionable insights . For example, using metrics to monitor readmission rates or treatment success helps hospital leadership identify areas for improvement. Such health performance metrics are essential for optimizing both patient outcomes and financial sustainability [1]. However, many large hospitals and research institutions have multiple BI teams focused on patient outcomes, another on billing, another on fundraising. These teams often need to collaborate, but they can't always share data freely. The reason? Patient-level data often contains personally identifiable information (PII) or protected health information (PHI), which is tightly regulated under laws like HIPAA and FERPA. This creates a problem when one team (e.g., the fundraising team) needs information owned by another (e.g., the clinical team), but accessing raw data directly would violate privacy policies. This is where aggregated metrics become essential. Instead of exposing individual records, teams can share high-level summaries like "average donation per patient type" or "total readmissions per quarter." These metrics are usually stored in aggregated metric tables, which represent rolled-up versions of raw data [4]. Instead of accessing sensitive patient-level records, teams can rely on these tables to answer important questions like, "What is the average emergency room wait time?" or "How many donations did each campaign receive last quarter?". However, even aggregated metrics can be risky. If a metric is based on a very small group of patients or includes fields like ZIP code or gender, it could still reveal sensitive information. To solve this, the paper proposes an novel AI-powered system that checks whether a metric is safe to share before it is used. This tool automatically analyzes the SQL query used to build the metric and evaluates whether it might accidentally reveal private data. It flags risky metrics that are too detailed or too narrow, and explains why they should be revised. The goal is to help BI teams work together while protecting privacy and staying compliant with regulations. Unlike older approaches that wait until after data is accessed or published, this system works ahead of time, giving teams instant feedback and peace of mind. It's modular, meaning it can fit into existing BI processes, and it helps create a secure environment for collaboration especially important in healthcare, where data privacy and ethical AI use are top priorities.

## II. BACKGROUND AND RELATED WORK

### A. Metric Abstraction and Aggregation

Metric tables, also referred to as aggregated metric datasets, are essential components in modern business intelligence (BI) pipelines. These tables contain pre-calculated, high-level statistical summaries such as averages, counts, or rates derived from underlying raw datasets. Rather than storing individual transactions or patient-level data, metric tables capture key performance indicators (KPIs) and other decision-driving metrics in an abstracted format [4].

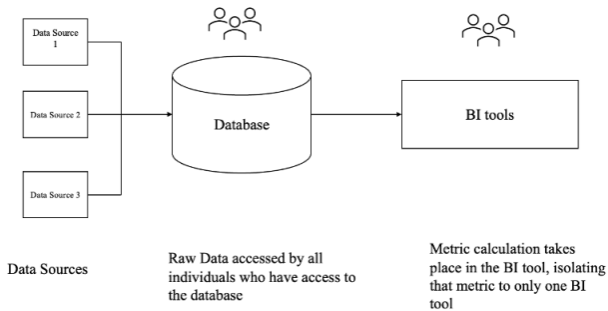

Fig. 1. Traditional BI architecture where raw data is directly accessed by multiple users, and metrics are computed within isolated BI tools.

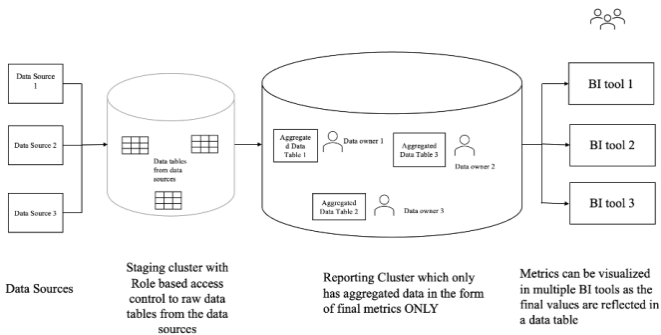

Fig. 2. Proposed architecture using aggregated metric tables and reporting clusters with role-based access, supporting secure metric sharing.

In traditional data architectures, raw data is often shared directly between teams, as illustrated in Figure 1. While this model facilitates access to granular information, it significantly increases the risk of exposing sensitive or personally identifiable information (PII), particularly in regulated domains such as healthcare or finance. When multiple teams consume the same raw datasets, inconsistencies in interpretation and logic, as well as unintended privacy violations, can arise.

To address these challenges, a more secure and governed architecture leverages aggregated metric tables, as shown in Figure 2. In this model, only the metric-owning team maintains access to the underlying raw data and is solely responsible for defining the metric logic. These finalized metrics, pre-aggregated, anonymized, and validated, are then stored in a centralized metric table that can be accessed organization-wide. This architecture offers two key benefits. First, it minimizes privacy risks by preventing unnecessary exposure to raw data while still supporting analytical needs across departments. Second, it enforces metric consistency and lineage. Since only the owner team controls the logic and transformation rules, any updates or refinements made to the metric automatically propagate downstream to all consumers, ensuring synchronization and reducing the chances of metric drift. This approach strengthens governance, enhances compliance readiness, and supports scalable data collaboration across business units.

While metric tables improve governance, they are not inherently immune to privacy risks. A core concern lies in re-identification attacks, particularly in scenarios involving small-N aggregates (e.g., average age of patients in a rural ZIP code). Such summaries can reveal sensitive patterns when group sizes are small or categories are narrowly defined. Attribute inference is another risk, in which an attacker leverages a machine learning classifier to infer a target user's private attributes (e.g., location, sexual orientation, political view) from its public data (e.g., rating scores, page likes) [5]. Furthermore, statistical leakage can occur when changes in an individual record disproportionately impact the metric (e.g., mean income in a group of three). These risks are amplified when metrics are published externally or shared across loosely governed teams. The leaking of confidential data to unauthorized entities can result in various problems for organizations and individuals. Similar to trade secrets, health records and banking details, leaks can affect the privacy of patients, and the security of accounts [6] Therefore, while aggregation offers a layer of protection, it is not foolproof without careful design and validation mechanisms. Several privacy-preserving frameworks have been proposed to address the risks inherent in data sharing and aggregation. As Organization scale and grow, the need for using data for decision making increases. However, such disclosures of personal health information even within the organizations raise serious privacy concerns. To alleviate such concerns, it is possible to anonymize the data before disclosure. One popular anonymization approach is k-anonymity [7]. which ensures that each record is indistinguishable from at least k-1 others on selected quasi-identifiers. While k-anonymity is effective for tabular data publication, it is limited when applied to aggregate metrics, as it does not consider the effect of small group sizes on statistical summaries. Differential privacy is the standard privacy protection technology that provides rigorous privacy guarantees for various data. This survey summarizes and analyzes differential privacy solutions to protect unstructured data content before it is shared with untrusted parties [8] , it offers a more robust mathematical guarantee by introducing calibrated noise to outputs. Moreover, most legacy validation systems are static based on hand-written rules and thresholds for column inclusion or minimum counts. These lack adaptability to evolving metric definitions and do not generalize well across organizations. In contrast, the proposed AI-driven framework evaluates metric-level SQL queries dynamically, using semantic understanding and syntactic structure to estimate re-identification risk, filling a critical gap in current governance strategies.

### III. System Architecture Overview

The proposed system is designed to assess whether a metric-generating SQL query risks overexposing sensitive information, such as gender or ZIP code, particularly when aggregation sizes are small. The workflow comprises five main stages. This pipeline is tailored for deployment within metric abstraction layers of BI systems, especially in privacy-regulated environments like healthcare institutions:

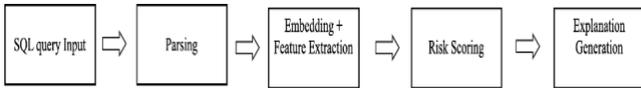

Fig. 3. High Level Workflow of the AI Framework

To understand how this AI system works, imagine it as a virtual data safety inspector that reviews SQL queries the instructions analysts write to pull data from a database. Before the system allows a query to run, it checks whether that query might accidentally reveal private or sensitive information (like gender, ZIP code, or birthdate) in ways that violate healthcare regulations like HIPAA. Let's walk through how each part of the system works behind the scenes using technologies that work similarly to language and pattern recognition tools in modern AI.

#### A. SQL Parser – Breaking the query into parts

When a human writes a SQL query, it might look like a complex sentence asking for something like: "Give me the average length of stay by gender and department." The parser is the first component that steps in. Think of it like how a grammar checker breaks a sentence into nouns, verbs, and objects. The SQL parser breaks the query into a structure called an Abstract Syntax Tree (AST). This tree helps the system "see" which columns are being used, how many tables are being joined, and whether groupings like gender or role are being used to summarize data. This gives the system a roadmap of what the query is trying to do, structurally.

#### B. CodeBERT – Semantic Representation

While the parser looks at how the query is structured, it doesn't always understand why the query was written that way. That's where CodeBERT comes in. CodeBERT is a special type of AI model trained on programming languages (including SQL) and natural language [9]. You can think of it like a translator that understands both the syntax and the intent behind code. It converts the SQL query into a numeric "fingerprint" that captures the underlying logic and purpose of the request. This is important because two different-looking queries can be trying to do the same risky thing, and CodeBERT helps catch those cases. For example, one query might group data by "zip," while another groups by a combination of "city + street" which can still lead to privacy risk. CodeBERT can sense that these are functionally similar even if they're written differently. By leveraging the pretrained transformer architecture, this model avoids the need for handcrafted semantic rules and instead relies on learned representations to distinguish queries that may pose privacy risks from those that do not.

#### C. Syntactic Feature Extractor – Pulling out risk signals

To make the system smarter, it doesn't just rely on language understanding. It also pulls out specific, hand-picked features that are known to raise red flags in data privacy. This includes:

- How many columns the query groups data by.
- How many tables it pulls data from (joins).
- Whether it uses sensitive columns like gender, date of birth, or ZIP code which, if used incorrectly, can make individuals identifiable.

These features act like "checkpoints" that further help the system understand the risk level of the query.

#### D. XGBoost Risk Classifier – Risk Classification

Once the system has the semantic fingerprint (from CodeBERT) and the hand-picked risk features (from the feature extractor), it combines them into a single profile of the query. This profile is then sent to XGBoost, a highly accurate machine learning algorithm that acts like a judge [10] . It has been trained on a large number of SQL examples labeled as safe or risky, so it knows how to spot dangerous patterns. The model outputs a risk probability score of 0 or 1, which reflects the likelihood that the query may violate privacy norms. We apply a configurable threshold typically 0.85 above which the query is automatically flagged as BLOCKED. This threshold can be tuned based on institutional policies, legal thresholds, or operational risk tolerance.

#### E. Explanation Engine – Telling you what went wrong

Blocking a query isn't helpful unless you explain why. So, the final part of the system is the explanation engine. This module reads the structure of the query (from the AST) and uses predefined templates to generate a human-readable explanation. For instance, it might say: "This query groups data by job title and shift, which may reveal individual identities because the group sizes could be too small." This is especially helpful for compliance teams or data stewards who need to fix the query without guessing what went wrong.

## IV. PSEUDOCODE AND ALGORITHM

This section outlines the pseudocode, giving a high level overview of the proposed AI framework. The full code is shared on github repository [11].

```
function detect_overexposure(sql_query):
    ast = SQLParser.parse(sql_query)
    embedding = CodeBERT.encode(sql_query)
    features = extract_syntactic_features(ast)
    combined_features = concatenate(embedding,
        features)
    risk_score = RiskClassifier.predict_proba(
        combined_features)
    if risk_score > 0.85:
        explanation = generate_explanation(ast
            , rules)
        return {
            "status": "BLOCKED",
            "risk_score": risk_score,
            "explanation": explanation
        }
    else:
        return {
            "status": "APPROVED",
            "risk_score": risk_score
        }
```

Listing 1. Pseudocode for SQL Risk Detection

## V. EXPERIMENTAL EVALUATION

This section evaluates the effectiveness of the proposed AI-driven SQL risk detection framework on a synthetic healthcare dataset. The evaluation simulates realistic business intelligence (BI) use cases in a hospital setting while assessing the system's ability to flag potentially privacy-violating SQL metric definitions.

### A. Dataset Description

A representative sample table named patient_data that mirrors anonymized hospital workforce and operational data is used for this evaluation. Due to the privacy regulations surrounding Protected Health Information (PHI) and Institutional Review Board (IRB) restrictions, real hospital data cannot be disclosed. As a result, we constructed a synthetic dataset that replicates the structure, schema, and statistical patterns of real-world healthcare data for the purpose of system validation. Each row represents an patient, and the dataset includes a blend of clinical and administrative attributes. The table contains the following fields:

TABLE I
SYNTHETIC DATASET SCHEMA AND COLUMN DESCRIPTION OF THE SAMPLE TABLE : PATIENT_DATA

| Column Name | Description |
| --- | --- |
| patient_id | Unique identifier for record traceability |
| dob | Sensitive |
| gender | Sensitive |
| zip | Sensitive |
| department | Medical unit |
| diagnosis_code | Medical condition classification: Sensitive |
| wait_time | Service-related metric |

For this experiment, dob, zip, gender, and diagnosis code are considered sensitive fields, as they may lead to re-identification or attribute inference when improperly grouped. The full synthetic dataset used for this experiment is available on the github repository [11].

### B. Query Evaluation Framework

A set of 3 SQL queries are generated representing a spectrum of aggregation risks from clearly benign to clearly problematic. The AI system parses each query using sqlglot, encodes the query using CodeBERT, and evaluates structural features such as number of GROUP BY fields, join count, and presence of sensitive columns. These features are fused into a vector and scored by an XGBoost classifier trained on labeled query examples. Any query with a risk score above 0.85 is flagged as BLOCKED and returned with a model-generated explanation. Below, is the model output on the 3 representative queries.

SQL QUERY 1 (Medium Risk): SELECT zip, COUNT(*) FROM patient_data GROUP BY zip;
AST NODES: ['GROUP BY', 'JOIN: 0', 'Columns: zip']
Risk Score: 0.87 Status: BLOCKED Explanation: ZIP code is a quasi-identifier and may expose individual locations if group size is small.

SQL QUERY 2 (High Risk): SELECT gender, diagnosis code, COUNT(*) FROM patient_data GROUP BY gender, diagnosis_code;
AST NODES: ['GROUP BY', 'JOIN: 0', 'Columns: gender, diagnosis_code'] Risk Score: 0.93 Status: BLOCKED Explanation: Grouping by gender and medical code may leak health-related sensitive segments.

SQL QUERY 3 (Moderate Risk): SELECT gender, COUNT(*) FROM patient_data GROUP BY gender;
AST NODES: ['GROUP BY', 'JOIN: 0', 'Columns: gender'] Risk Score: 0.74 Status: APPROVED Explanation: While gender is sensitive, group sizes are likely sufficient to prevent leakage.

### C. Comparison to Rule-Based Systems

Traditional rule-based approaches to query risk detection typically rely on static filters or hard-coded blacklists such as banning combinations like "zip" and "dob" within the same GROUP BY clause. While straightforward, these systems often suffer from three key limitations. First, they lack context-awareness; for example, grouping by gender alone may be perfectly acceptable in many cases, yet a rule engine might flag it indiscriminately. Second, such systems are rigid and struggle to adapt to novel data patterns or edge cases that emerge in evolving BI environments. Third, they offer no mechanism for prioritization or explanation producing binary outcomes without confidence scores or rationale, which hinders transparency and governance. In contrast, the proposed AI-powered system addresses these challenges through dynamic, data-informed evaluation. It computes a continuous risk score from 0 to 1, enabling more nuanced decision-making. For example, Query 3 in the evaluation includes a sensitive field but was assessed by the AI as posing minimal actual privacy risk, given its structural context and the distribution patterns typically found in hospital datasets. A traditional rule engine would have flagged it regardless. Moreover, this system provides clear, interpretable explanations for each decision whether a query is approved or blocked. These explanations enhance audit readiness, enabling data stewards, BI analysts, and compliance officers to understand not only what decision was made, but why. This fosters trust, supports IRB and HIPAA compliance, and encourages proactive data governance in healthcare settings.

In summary the proposed model demonstrated strong domain-aligned sensitivity. High-risk queries were accurately flagged based on grouping patterns and the presence of identifiable fields. Low-risk queries that provide analytical value without privacy compromise were rightfully approved, supporting operational continuity.

## VI. Limitations and Assumptions

- Keyword-Based Sensitive Field Detection: The current AI framework identifies sensitive fields such as dob, gender, zip, and role using a fixed list of keyword matches. While effective for standard naming conventions, this approach lacks adaptability to schema variations or domain-specific terminology. For instance, fields labeled as date_of_birth or postal_code may not be detected. To enhance robustness, future iterations should support configurable sensitive term lists and semantic similarity techniques (e.g., vector embeddings) to identify synonymous attributes across diverse datasets.
- Contextual Risk in JOINs and GROUP BY Clauses: The model evaluates structural query risks using features such as the number of JOIN operations and the presence of GROUP BY clauses. However, true re-identification risk is highly context-dependent, especially influenced by the cardinality of grouped results. Because the current framework performs only static query analysis (without executing the query), it cannot assess group sizes or distribution patterns. Future versions could incorporate metadata-driven heuristics (e.g., value distribution histograms or minimum subgroup thresholds) to better approximate privacy risk at design time.
- Handling of Advanced SQL Constructs: While the current implementation successfully parses and evaluates standard SQL queries, it may exhibit degraded performance when encountering advanced constructs such as Common Table Expressions (CTEs), window functions, and deeply nested subqueries. These complex patterns may yield incomplete or shallow Abstract Syntax Trees (ASTs), limiting downstream feature extraction. Enhancing the parser to recursively analyze these structures or pretraining the model on a wider set of syntactic patterns would improve its generalization to production-grade BI environments.
- Model Bias :Additionally, the training data and labeling process used to build the classifier may carry institutional biases reflective of the source environment. Broader calibration across multiple healthcare systems would improve fairness.

## VII. Conclusion

In this paper, an AI-assisted framework is introduced that supports metric abstraction as a privacy-preserving alternative to sharing raw data across teams in healthcare organizations. As hospitals and research institutions rely more heavily on analytics for operational and clinical decision-making, the need to collaborate without compromising sensitive information like patient demographics or diagnosis codes has never been more urgent. This framework provides a technical solution that enables this balance. Instead of relying on runtime data access or manual reviews, the system analyzes SQL-based metric definitions before they are executed. By parsing the structure of each query and combining it with semantic embeddings (via CodeBERT) and syntactic features (e.g., number of joins, presence of sensitive fields like ZIP or DOB), the model predicts a privacy risk score using a trained XGBoost classifier. Queries flagged as high risk are blocked and accompanied by a clear explanation, such as: "Metric groups by job_title and shift, which may reduce group sizes below FTC thresholds." This approach allows organizations to safely publish aggregate metrics like readmission rates or department-level wait times without exposing raw patient records. It also ensures that BI teams outside the data-owning department (e.g., fundraising or operations) can access needed insights without direct access to sensitive tables. The modularity of the system makes it easy to extend: sensitive keywords can be configured, more advanced SQL constructs can be added to the parser, and the model can be re-trained on domain-specific query patterns. Compared to static rule-based filters, this framework brings flexibility and context-awareness to the privacy review process. It supports pre-execution validation, meaning risks are caught before any query runs against protected data. The result is a governance tool that's not only compliant with HIPAA, IRB, and zero-trust principles but also practical for real-world BI workflows. Overall, this system lays the groundwork for ethical AI in metric generation and collaborative data infrastructure. It enables organizations to move beyond "access or deny" models and instead adopt intelligent, explainable privacy controls at the metric definition layer. For institutions navigating the tension between data utility and privacy compliance, this framework offers a path forward that is both scalable and responsible.